\documentclass[a4paper,fleqn]{cas-dc}

\usepackage[authoryear,longnamesfirst]{natbib}
\usepackage{import}
\usepackage{hyperref}
\usepackage{float}
\usepackage{verbatim} 
\usepackage{apalike}
\restylefloat{figure}
\restylefloat{table}
\usepackage{caption}
\usepackage{import}
\usepackage{xcolor}
\usepackage{soul}
\usepackage{graphics}
\usepackage{algorithm}
\usepackage{algpseudocode}
\usepackage{url}

\def\tsc#1{\csdef{#1}{\textsc{\lowercase{#1}}\xspace}}
\tsc{WGM}
\tsc{QE}
\tsc{EP}
\tsc{PMS}
\tsc{BEC}
\tsc{DE}

\begin{document}
\let\WriteBookmarks\relax
\def\floatpagepagefraction{1}
\def\textpagefraction{.001}

\shorttitle{Modified Query Expansion Through Generative Adversarial Networks}

\shortauthors{Cakir A., Gurkan M.}

\title [mode = title]{Modified Query Expansion Through Generative Adversarial Networks for Information Extraction in E-Commerce}                      


%
\author{Altan Cakir}[type=editor,
                        auid=000,bioid=1,
                        orcid=0000-0002-8627-7689
                        ]

\cormark[1]
\fnmark[1]
\credit{Conceptualization of this study, Methodology, Software}



\author{Mert Gurkan}
\fnmark[2]


\credit{Data curation, Writing - Original draft preparation}

\cortext[cor1]{Corresponding author}

\fntext[fn1]{Physics Engineering, Faculty of Science and Letters, Istanbul Technical University, Istanbul, Turkey and Istanbul Technical University Artificial Intelligence, Data Science Research and Application Center, Istanbul Turkey}
\fntext[fn2]{Insider (useinsider.com), Istanbul, Turkey}


\begin{abstract}
This work addresses an alternative approach for query expansion (QE) using a generative adversarial network (GAN) to enhance the effectiveness of information search in e-commerce. We propose a modified QE conditional GAN (\textit{m}QE-CGAN) framework, which resolves keywords by expanding the query with a synthetically generated query that proposes semantic information from text input. We train a sequence-to-sequence transformer model as the generator to produce keywords and use a recurrent neural network model as the discriminator to classify an adversarial output with the generator. With the \textit{modified} CGAN framework, various forms of semantic insights gathered from the query-document corpus are introduced to the generation process. We leverage these insights as conditions for the generator model and discuss their effectiveness for the query expansion task. Our experiments demonstrate that the utilization of condition structures within the \textit{m}QE-CGAN framework can increase the semantic similarity between generated sequences and reference documents up to nearly 10\% compared to baseline models.
\end{abstract}

\begin{keywords}
Generative Adversarial Networks \sep Query Expansion \sep Conditional Neural Networks \sep Information Retrieval \sep E-Commerce
\end{keywords}



\maketitle

\section{Introduction}
In search based business models, such as e-commerce, given a search query, the system needs to match it to some relevant keywords/categories/frequencies by business partners and then pull out the related category/product for query searching and ranking. The query keyword matching can be done by some simple matching rules like exact match, similarity match, and phrase match, which are all based on matching the similar tokens shared by query and keywords. On the other hand, using AI-based recent techniques for smart match is an important yet difficult match type that can associate a query to some relevant keywords even they do not generate many similar tokens.

In general, it is well defined that search queries mathematically follow the power law distribution \citep{powerlaw}. The curve formed by the most frequent queries constitutes the main center, while the rare queries with low frequency form the tail of the curve. Although they are few in such cases, low-frequency queries are excluded from the query volume traffic as a whole and therefore cause problems in systems as a data deficiency that needs to be generated synthetically.

Because of the incoming query distribution to the search engine, the performance of matching rare queries with documents existing in the database is a challenging task. It is often the case that an additional process is required to assist the match between rare queries and documents. To address the problem, various methodologies such as relevance feedback methods, similarity-based methods for query-document matching, machine translation models for query transformation, and query expansion methods are discussed in the literature. 

Query expansion is one of the significant problems studied in the Information Retrieval (IR) domain with various applications such as question answering, information filtering, or multimedia document matching tasks \citep{survey2012}. The problem can be described as the attempt of the increasing performance of matching input sequences and document the corpus of an IR system by reformulating given input sequences \citep{Azad_2019}. 
Query expansion methodologies are often applied where the input queries are words or sequences originating from real human users, while documents to match or rank them consist of predefined items. Natural language queries that match to same documents can differ verbally and semantically \citep{furnas}. Because of this ambiguity, the complexity of query-document matching is often increased by the innate characteristics of the data.

Earlier studies in the query expansion domain seem to focus on rule-based applications. These applications evaluate candidate expansion terms by the frequency of appearing together with the words in the original query \citep{carpineto2001}. In addition to word frequency based studies, systems built upon pseudo-relevance feedback structures are also widely utilized in the literature. \citep{rm3} uses the Markov random fields for modelling dependencies to assist the query expansion process. \citep{tensorr} provides a different approach to the query expansion methods with pseudo-relevance feedback, where they build tensor representations of queries that enables obtaining relevance feedback based on word meanings. 

Adoption of the deep learning applications in the natural language domains generated word embeddings as efficient ways to represent semantic information of text data \citep{word2vec}. The utilization of word embeddings made it possible to evaluate the semantic relationship between words. This capability is employed for query expansion problems by using various ways to evaluate the similarity of words that make up the queries and candidate terms to expand these queries.  

The popularity of the word embedding methods for various problems for IR and NLP, led research efforts to increase the accuracy of word representations in specific cases. To this end, alternative ways to produce different embeddings of tokens for query expansions are proposed \citep{Sordoni_Bengio_Nie_2014}. Additionally, research conducted utilization of task-specific trained word embeddings for query expansion \citep{localembed}. This way, word representations are more likely to capture the context and semantic properties of the trained corpus. 
Following these works, \citep{prophetnetads, lian2021end} proposed a query expansion approach for search engine optimization by utilizing a prefix tree to serve as look ahead strategy for generating expansion terms for given queries. 

Recent applications of GAN methods provide alternative methods to approach the problem. GAN models can directly generate expansion terms or expanded user queries by training over user search queries and their matching documents. In GANs the discriminative network can learn to distinguish between the synthetic data created by the generator and the real data examples. This way, the generation process is challenged by the network itself to create high-quality samples. This approach of training has proven to be very successful in the computer vision domain and increasing its popularity in natural language processing problems. Additionally, the research focusing on establishing back-propagation between discriminator and generator models with discrete tokens in text data \citep{yu2017seqgan, gumbel} provided highly performing generative models.  

With initial GAN models, the model is trained with noise for the generation process. With conditional structures, the query generation of the GAN models can be assisted with the chosen condition mechanism. Similar to earlier works in the query expansion domain, enhancing user queries with existing relevant information is adopted by GAN-based architectures too. GAN models can utilize part of text data, class labels present during the training, or extracted properties of the query and documents as conditions to increase the likelihood of matching queries with desired documents. The study of \citep{lee2018rare} proposes a conditional GAN structure with a query expansion approach for enriching rare queries in search engines. The study of \citep{huang2021gqe} employs a well-known method of pseudo-relevance feedback in the query expansion domain as the condition for their expansion term generation. 

Studies discussed intend to create a conditional GAN-based framework to leverage query expansion to match keywords for an effective search selection. In general, a sequence-to-sequence model, in which the input sequence is a random word vector followed by a query vector, is commonly used for the generator. The output sequence composes of the vectors of the generated keywords. As the discriminator, the parallelized Recurrent Neural Network (RNN) model is used as a binary classifier. However, most of these studies are not conducted from the perspective of improving search engines by enhancing query-document matching performance. Our study aims to combine GAN architecture and existing query-enhancing methods by utilizing them as condition structures for the generator model. Proposed conditional GAN models aim to alleviate the performance drop of search engines, by increasing the query-document matching performance with condition-assisted query expansion mechanisms. 

To alleviate the effects of the problem described, we introduce the \textit{m}QE-CGAN (Modified Query Expansion Conditional Generative Adversarial Network) framework to study the query expansion to enhance the performance of a search engine by increasing the query-document matching performance. The generator of the model is a sequence-to-sequence encoder-decoder model that takes user search queries and the vectors from the applied condition mechanism. The output of the generator, expanded queries, is evaluated by the discriminator model. We use an LSTM model for the binary classification task between the synthetic and real samples. During adversarial learning, the evaluation of the discriminator guides the performance of the generator model. With the \textit{m}QE-CGAN framework, our contributions can be listed below;

\begin{itemize}
    \item \textbf{Model:} We propose a novel conditional generative adversarial network model that takes the semantic relationship between the query and document pairs as conditions. The generator of the model is a sequence-to-sequence encoder decoder model, while the discriminator is an LSTM-based binary classifier. We provide details of the model framework and the evaluation of the training process with a conditional approach.
    \item \textbf{Conditional Query Expansion:} We provide alternative methods for condition structures with generative adversarial networks. Condition structures discussed in this paper aim to capture semantic relationships between query-document pairs.
    \item \textbf{Datasets:} We test our generative model with the user query and document pairs from the customers of Insider\footnote[1]{\href{https://useinsider.com/}{https://useinsider.com}}. By testing the proposed models with different customer datasets, we evaluate our models against data with different characteristics.
\end{itemize}

The primary aspect that \textit{m}QE-CGAN framework differs from the existing conditional GAN frameworks is that the models of the framework are conditioned on the semantic and statistical relationships between the query-document data. Employed conditions are not limited to the individual relationships between the query and the matching document pairs. They are rather constructed with the consideration of the entire corpus. Hence, the generation process of the GAN framework utilizes conditions produced after the semantic analysis of the entire corpus. 

\section{System Architecture}
\subsection{\textit{m}QE-CGAN Framework}

The proposed framework for adversarial training with the \textit{m}QE-CGAN framework can be observed in the Figure \ref{fig:framework} below. The generator model of the framework takes input queries and the selected condition vectors assigned for input queries. With a sequence-to-sequence structure, it generates expansion terms from the given queries. The discriminator model of the adversarial schema performs binary classification on the expanded synthetic queries and documents that match to original queries of users.

Condition generation mechanisms discussed in the study aim to take advantage of the data used. As the query-document pairs in datasets denote user searches and matching documents, condition approaches focus on the semantic and similarity metrics of given queries and their matching documents by the search engine.

\begin{figure*}[ht]
\centering
\includegraphics[width=1\textwidth]{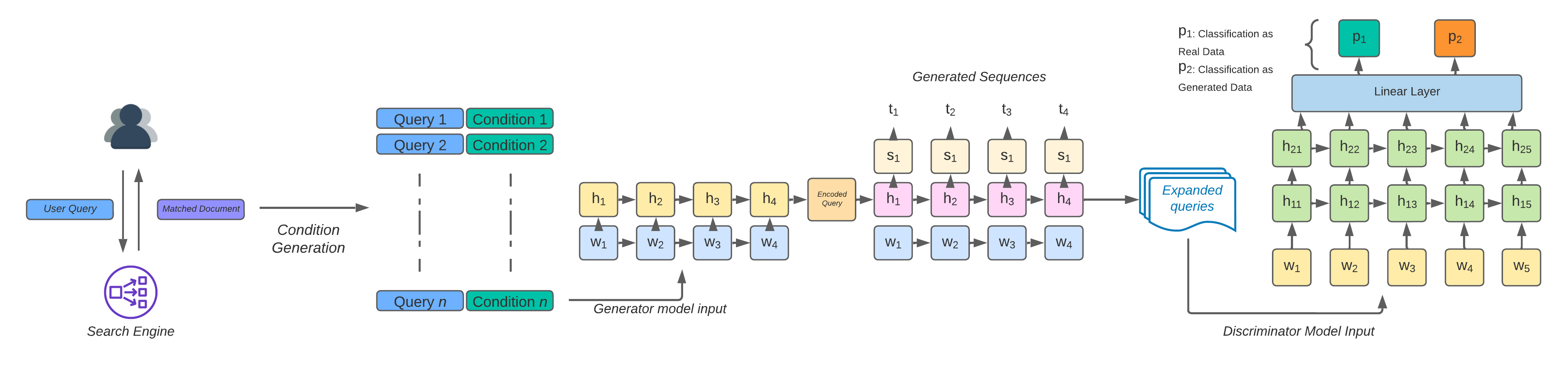}
\caption{Diagram of the \textit{m}QE-CGAN framework.}
\label{fig:framework}
\end{figure*}

\subsubsection{Generator Model}
The generator model of the architecture is an encoder-decoder sequence-to-sequence model that takes FastText \citep{bojanowski2016enriching} word embedding representations of the user search queries and their corresponding condition vectors as input. To be able to achieve back-propagation with the discrete input sequences, similar to the existing studies \citep{yu2017seqgan, lee2018rare} Monte Carlo rollouts are used in the decoder of the generator. With this method, rewards produced by the discriminator can be transferred to the generator for each generation step.

\begin{figure*}[ht]
\centering
\includegraphics[width=0.8\textwidth]{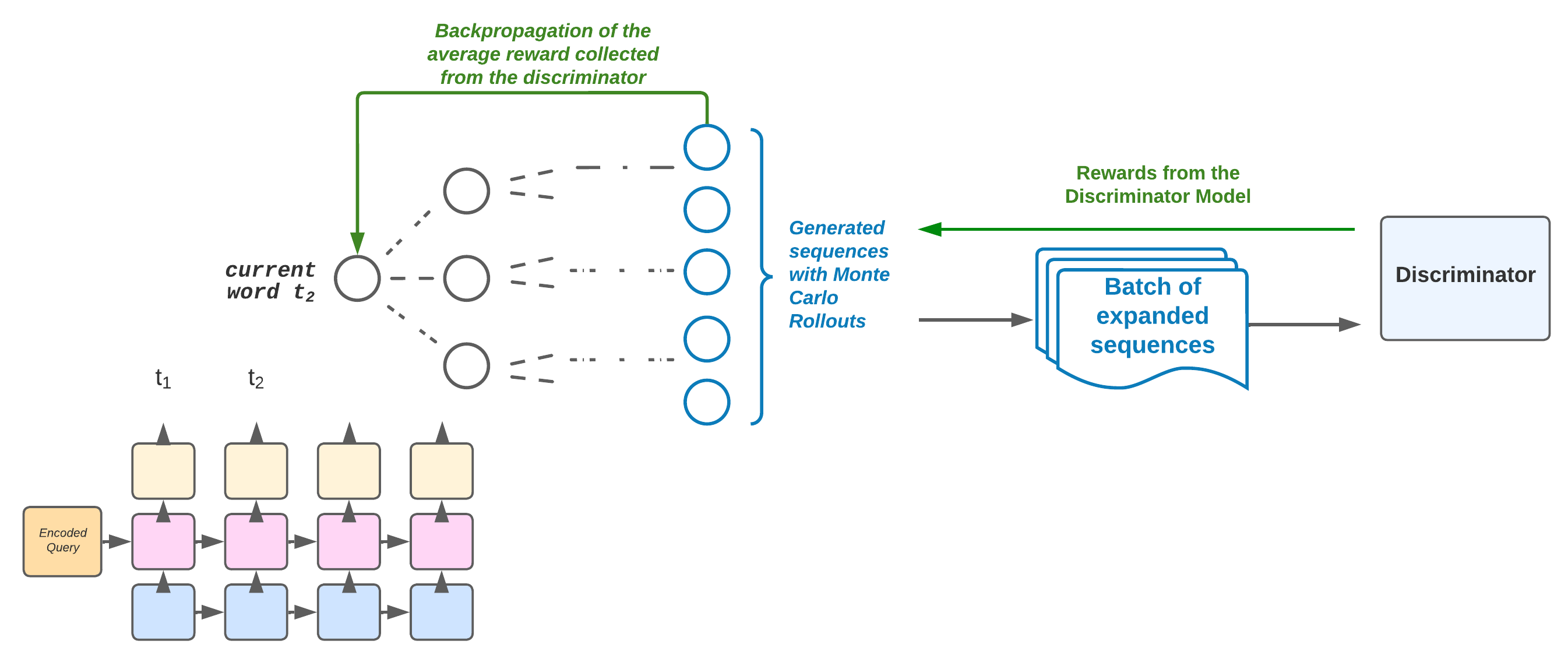}
\caption{Diagram of the Monte Carlo rollouts. At each step, a batch of sequences are generated by the decoder of the network. These batches are evaluated by the discriminator to guide the generation process of the generator model.}
\label{fig:framework}
\end{figure*}

\subsubsection{Condition Structures}
GAN models can be extended into conditional models if the adversarial learning process is performed with additional information \citep{cgan}. With the introduction of conditions, the models can be inclined to generate samples with the desired qualities \citep{NIPS2015_8d55a249}. Conditions are introduced to guide the generator model during the sequence generation process. Condition structures utilized in this study are generated before training the model. Condition vectors of queries are concatenated with the word embedding representation of the user queries during training. To retrieve them, Ball Tree-based look-up tables are used. 

To this end, four different condition structures are applied with the following expected priorities; (1) It should enrich the user query with other similar queries, and (2) it should provide information that will assist in distinction between similar documents that can be mapped with the given query. To address these requirements, various condition vector generation strategies displayed are implemented. Utilized methods are considered to be addressing the shortcomings of the encoder-decoder generator model. These condition generation strategies are described in the list below.


\begin{enumerate}
    \item Query Weighting with TF-IDF Scores: Condition vectors are generated with CBOW representations of the TF-IDF weighed input word embeddings
    \item Search Tree Based Document Similarity: Condition vectors are generated with CBOW representations of the most similar documents to the given input query.
    \item Search Tree Based Word Similarity: Condition vectors are generated with CBOW representations of the most similar words in the corpus of documents to the given input query.
\end{enumerate}

Although these methods are commonly utilized in query expansion approaches \citep{azadanew}, their integration as condition mechanisms is not adequately experimented with generative models.

\subsubsection{Discriminator Model}
The discriminator model of the \textit{m}QE-CGAN framework is built with the same pre-trained Fasttext word embeddings and LSTM layers processing embedded representations of generated and real document sequences. Unlike the generator model of the framework, the discriminator model does not utilize the condition structures for its pre-training and adversarial learning processes. The model is designed for the binary classification task between real documents in corpus and sequences formed by the generator as the synthetic data. 

\begin{figure}[H]
\centering
\includegraphics[width=0.35\textwidth]{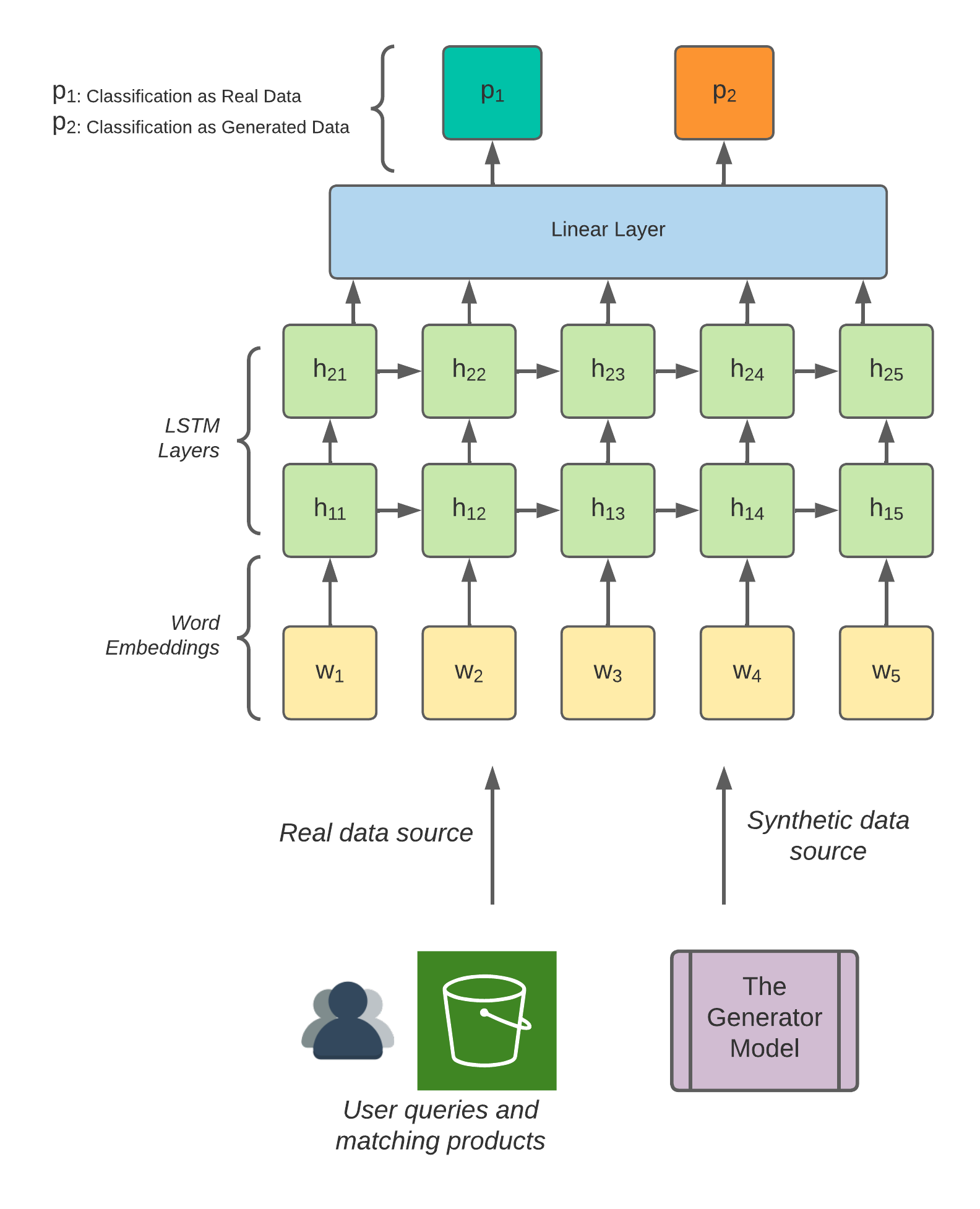}
\caption{LSTM based discriminator model of the \textit{m}QE-CGAN framework.}
\label{fig:disc}
\end{figure}

\subsection{Implementation Details}

We conducted the implementation with the PyTorch library \citep{NEURIPS2019_9015} in this study. The encoder-decoder generator model is implemented by using the \textit{TransformerEncoder} and \textit{TransformerDecoder} classes in PyTorch. The generator model uses 2 layers for both the encoder and the decoder parts. Initially, the input user queries are transformed to FastText word embedding representations with each word being represented with a tensor of size 100. Originally, FastText word embeddings are available for Turkish with a size of 300. To reduce the amount of GPU RAM required, we transformed these embedding representations to vectors with size 100 with the \textit{reduce\_model} implementation of the FastText library. It is followed by applying positional embedding to assign the order context to tokens in sequences with the help of the attention heads \citep{NIPS2017_3f5ee243}. For the forward pass, the given query and its paired condition vector are concatenated. The encoder and decoder of the generator take an input size of 200 from the concatenated tensors, and they have a hidden size of 512. 

Pre-training of the generator is performed by training the generator model with the learning rate $10^{-3}$ and the Adam \citep{adam} optimizer. During pre-training, the generator uses a softmax layer of size $N$, where $N$ is the total vocabulary size of the query and document corpus. Sequence generation is performed iteratively by predicting an expansion term at each step until the generator predicts the next token as $<EOS>$ (end of the sequence) token. For many cases, it was observed that after training the generator 16 epochs the Cross-Entropy Loss of the model does not improve. 

The discriminator model of the framework is intentionally kept simpler than the generator. For the discriminator, we used a 1 layer LSTM model. To decide on the hyper-parameters of the discriminator, a grid search is applied to hyper-parameters by training discriminator models with combined datasets of synthetic data from the generator and the samples from the document corpus. The discriminator model where the loss is optimized was obtained with the number of epochs as 24, the learning rate as $10^{-2}$, dropout as $0.1$, and the batch size as 256.

\section{Experiments}
\subsection{Datasets}

The datasets utilized in the study are generated by the analysis of user behavior in a search engine product of Insider. More specifically, these datasets consist of user search queries and the first-ranked resulting products in the platforms of Insider customers. It should be noted that the datasets utilized in this study do not include any specific user information. During the data collection step, any information that can be exploited to identify the user information is discarded. 

As the general user behavior in search engines is to enter fewer words to match the desired documents \citep{qucat}, queries in search engines tend to compose fewer words compared to the documents. This general observation is also present in the datasets utilized in our study. The average number of words in queries and documents in datasets used in the study can be observed in Figure \ref{fig:avglen} below. 

\begin{figure}[H]
\includegraphics[width=0.45\textwidth]{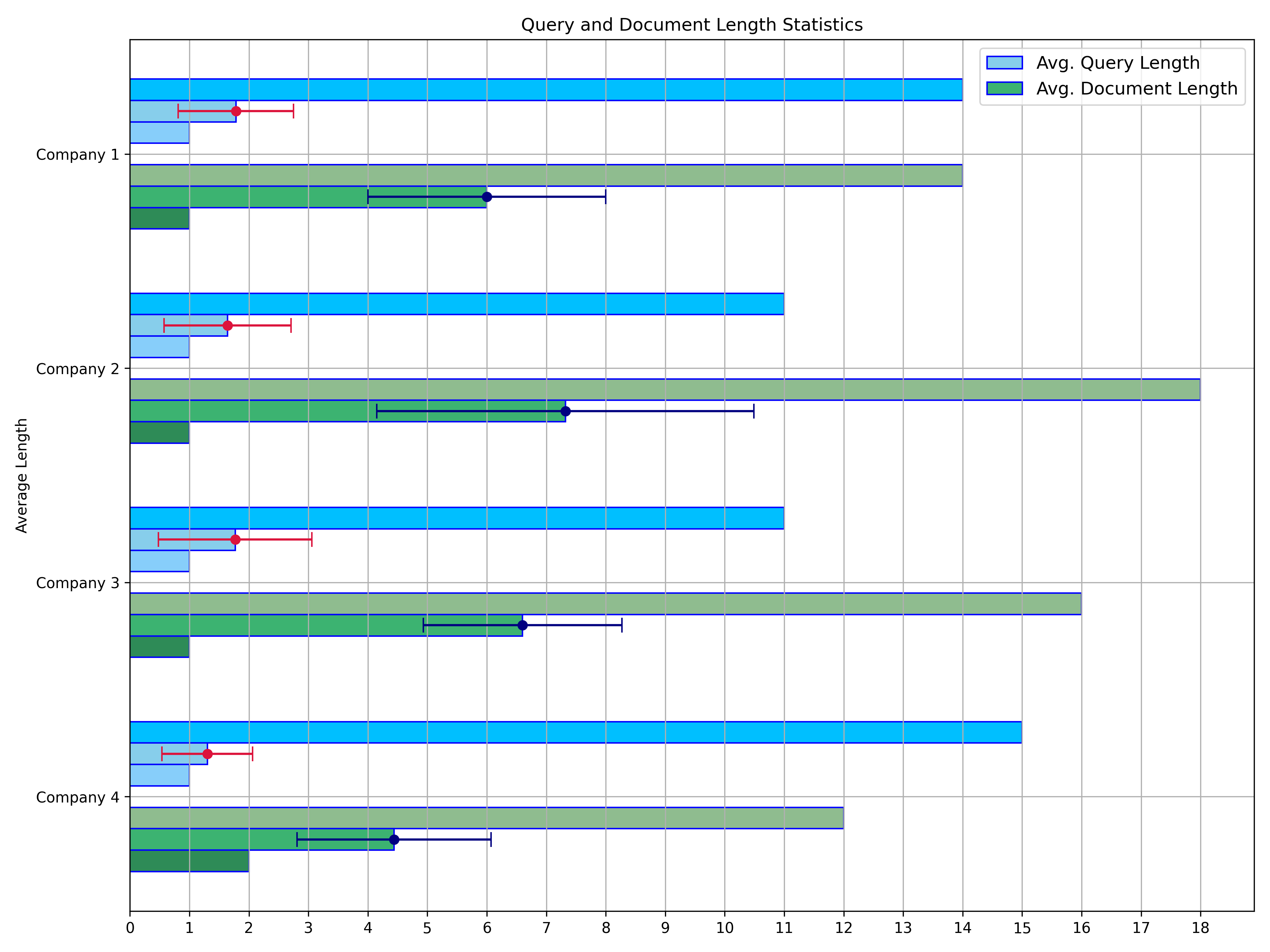}
\caption{Statistics of the query and the document datasets utilized in the study. For each dataset, bars at the top display the maximum, average, and minimum number of words in queries. Similarly, bottom bars display statistics of the document corpus. For all datasets, the average number of words in user searches are almost four times less than their matching product equivalents, suggesting further ways to employ semantic information to be extracted from document data.}
\label{fig:avglen}
\end{figure}

The difference between the number of words in queries and documents introduces various challenges for search engines. In the case of rare query inputs of users, similar to recommendation systems search engines are more prone to the cold start problem \citep{camacho2018social}. Datasets generated in the study aim to challenge the \textit{m}QE-CGAN framework in this regard. 

\subsection{Experimented Evaluation Metrics}

Both the generator and the discriminator of the \textit{m}QE-CGAN framework are trained with cross-entropy loss during pre-training processes. For model comparisons, changes in the perplexity metric were analyzed for the generator. For the discriminator, the accuracy of the trained models was tracked.

In addition to these metrics, we track the language diversity of the expanded queries. To this end, a new evaluation metric, the Word Coverage (WC), is defined. Word Coverage metric checks the ratio of the number of unique words selected as expansion terms by the generator to the number of unique words in the document corpus. For a successful model, we expect this metric to be close to 1. Obtaining a Word Coverage metric lower than one suggests that the generator model was not able to cover words in the tested set in the query expansion process. On the other hand, obtaining a Word Coverage metric higher than one indicates that the word selection process during query expansion utilized more unique words from the training corpus than it should have. The formula of the metric can be observed below. In the formula, $s_{QE}$ denotes words that are selected as expansion terms by the generator, $s_{C}$ denotes the words in the tested corpus.

\[ WC = \frac{\sum uniq(s_{QE})}{\sum uniq(s_{C})} \]

In addition to analyzing the expansion term diversity in generated sequences, models are also evaluated by the semantic similarity between generated sequences and reference sequences. To this end, we utilized average cosine similarity between the generated sequences obtained with expansion terms and their corresponding references in the document corpus. To assess the similarity, the average CBOW representations of both sets are compared. CBOW representations are obtained by averaging the embedding representations of the words that make generated and reference sequences. The formula below summarizes the Semantic Similarity (SS) analysis between generated and reference documents.

\[ SS =   \sum_{i}^{N}\frac{\hat{w}_i \cdot w_i}{\left\| \hat{w}_i\right\| _{2}\left\| w_i\right\| _{2}}  \]

These metrics allow us to assess the success of generated sequences without penalizing the n-gram matching performance of the generator. As the significance of n-gram matching and the word order are less crucial for matching user queries and products, the metric provides significant insights into the generation performance with different datasets. 

\subsection{Generator Evaluation Metrics}

Resulting evaluation metrics after integrating the condition generation strategies to the generator model can be found from the table below.

\begin{table}[H]
    \captionsetup{font=scriptsize}
    \centering
    \resizebox{1\linewidth}{!}{
    \begin{tabular}{ c|c|c|c|c|c } 
    \hline
    Dataset & Condition & CE Loss & Perplexity & WC & SS ($\mu$, $\epsilon$)\\
    \hline
    \hline
    \multirow{4}{4 em}{C1} & Baseline Generator & 1.266 & 3.650 & 1.07 & (0.602, 0.173) \\ 
    & Word Sim. & 1.328 & 3.792 & 1.02 & \textbf{(0.696, 0.169)}\\ 
    & Document Sim. & 1.258 & 3.536 & 0.99 & (0.659, 0.178) \\ 
    & TF-IDF & 1.288 & 3.644 & 1.15 & (0.606, 0.176) \\ 
    \hline
    \multirow{4}{4 em}{C2} & Baseline Generator & 0.267 & 1.307 & 0.46 & (0.898, 0.144) \\ 
    & Word Sim. & 0.27 & 1.311 & 0.45 & \textbf{(0.911, 0.14)} \\ 
    & Document Sim. & 0.272 & 1.313 & 0.46 & (0.902, 0.1412) \\ 
    & TF-IDF & 0.267 & 1.307 & 0.46 & (0.894, 0.146) \\ 
    \hline
    \multirow{4}{4 em}{C3} & Baseline Generator & 0.34 & 1.405 & 1.07 & (0.662, 0.173) \\ 
    & Word Sim. & 0.337 & 1.401 & 0.84 & (0.81, 0.169) \\ 
    & Document Sim. & 0.344 & 1.411 & 0.98 & (0.809, 0.171) \\ 
    & TF-IDF & 0.33 & 1.391 & 0.74 & \textbf{(0.819, 0.162)} \\  
    \hline
    \multirow{4}{4 em}{C4} & Baseline Generator & 1.292 & 3.650 & 1.26 & (0.709, 0.217)  \\ 
    & Word Sim. & 1.285 & 3.626 & 1.02 & \textbf{(0.736, 0.209)} \\ 
    & Document Sim. & 1.28 & 3.605 & 1.15 & (0.721, 0.203) \\ 
    & TF-IDF & 1.218 & 3.39 & 1.20 & (0.686, 0.272) \\  
    \hline
    \end{tabular}
    \caption{Generator evaluation metrics of the selected dataset of companies. Company names are replaced with placeholders as C. To provide further context; Company 1 (C1) is a Turkey-based cosmetics company, Company 2 (C2) and 4 (C4) are fashion retailers originated in Turkey, and Company 3 (C3) is a worldwide technology company.}
    \label{tab:my_label}}
\end{table}

In Table \ref{tab:my_label} above, the Baseline Generator is trained by self-conditioning the input user queries. This way, the effectiveness of the condition structures is evaluated against a condition mechanism that will not provide further positive cues for the generation process. WC denotes the Word Coverage metric discussed earlier, and SS denotes the Semantic Similarity metrics of trained generators. The mean and standard deviation of cosine similarities between generated sequences and reference documents can be observed in the table. Although generators with different models yield similar Cross Entropy Loss values, the Semantic Similarity obtained from generators with word similarity as conditions result in more successful generation processes. The Word Coverage metric is higher than it should have been for baseline generator models, compared to models trained with additional conditions.

These metrics were obtained after training each generator model 16 epochs with the Cross-Entropy Loss function. As our initial observations demonstrated that generator pre-training tends to not improve after 16 epochs, Table \ref{tab:my_label} displays the effectiveness of condition methods before adversarial learning. 

\subsection{Adversarial Learning}

For adversarial learning, we pre-trained the generator and the discriminator models with half the number of epochs mentioned in the \textit{Implementation Details} section. Thus, these models were not optimized for the underlying dataset. The pre-training of the generator is performed with the train and validation splits, where the discriminator is trained with the test splits. Below, the adversarial learning algorithm we use with these configurations can be observed.

\begin{algorithm}[h]
\caption{Adversarial Learning with Policy Gradients}\label{alg:cap}
\begin{algorithmic}
\Require Generator pre-training policy $G$; rollout policy $G_{r}$; Discriminator pre-training policy $D$; query-document dataset $S = \{X_{1:N}, Y_{1:N}\}$
\State Pre-train G using Cross-Entropy Loss on $S$
\State Generate synthetic examples using G for training D as $S_{\theta}$ 
\State Pre-train D using Cross-Entropy Loss on $\{S, S_{\theta}\}$
\Repeat
    \For{\texttt{e in epochs}}
        \For{\texttt{b in batches}}
            \State Generate $N$ rollout sequences with $G_{r}$ for $S_b$ 
            \State Obtain average reward from $D$ as $R_b$
        \EndFor
        \State Update $G_{r}$ with $avg(R_b)$
    \EndFor
    
\Until G loss of \textit{m}QE-CGAN does not improve
\end{algorithmic}
\end{algorithm}

During adversarial learning, at each expansion term generation step the generator model samples $N$ finished sequences from unfinished sequences with Monte Carlo rollouts. These sampled sequences are evaluated by the discriminator $D$ to inform the generator model $G_r$ about the current generation step. The average discriminator loss $avg(R_b)$ obtained for this operation is used for rewarding the generator model and updating its parameters. These operations are repeated for each batch in the query-document dataset. By employing Policy Gradients \citep{policy_grad}, we convert the discriminator loss to the format that the generator can utilize. 

\section{Results \& Discussion}
Table \ref{tab:my_label} demonstrates that the generator models conditioned with the Word Similarity method result in the best semantic evaluation metrics. Word Similarity provides precise embedding vectors of words that are the most similar in meaning to the words in the query. In this manner, models conditioned with it receive more insights about the context of the given query. The approach can also be considered similar to the pseudo-relevance feedback methods where the query is enhanced with the documents that initially matched with them. Compared to Word Similarity conditions, Document Similarity and TF-IDF Weighting conditions yield slightly worse semantic evaluation metrics. In some cases, condition methods do not increase the generator model performance compared to the baseline model. Our analysis displayed that Document Similarity conditions tend to guide the generation process in inaccurate ways, as the most similar documents to given input queries were possible to be differentiating from the reference documents. For many cases, TF-IDF Weighting seems to omit words in a given query to incline the generator model to narrow the space for expansion term selection. 

When model performances are compared among datasets, it can be seen that the models were most successful in the Semantic Similarity metric for the dataset of Company 2 (C2 in Table \ref{tab:my_label}) and least successful for the dataset of Company 1 (C1). This result was expected after we analyzed the properties of different datasets in the study. The C1 dataset is the most challenging dataset having the largest vocabulary size among utilized datasets. On the other hand, the C2 dataset can be considered more trivial among others as having the smallest vocabulary size and documents are composed of more keywords. The evaluation metrics of conditioned models tested with the C3 dataset should also be noted. As it is a dataset of a technology company, product name documents are often composed of words that do not have much meaning and context when separate. Here, prioritizing the words with higher TF-IDF scores for the generator model can increase the generator performance to select more precise expansion terms within the same query context.

When the extended queries produced by the models are examined, it is seen that the expansion terms in the generated sequences have a high success in being in the same context as the reference document, but the prediction of the terms in the documents is not at the same level. It should also be noted that the datasets used for training the models are limited. When the system we proposed in our study works integrated with a search engine, it will be better optimized with real-time data flow with higher traffic. We expect the success of word prediction in sequences to increase even more. Furthermore, due to the nature of the problem we aim to address, it is more important that the added words bring the semantic values of the extended queries closer to the documents instead of directly matching the words added in the expanded queries with the documents tested. Therefore, evaluation metrics such as BLEU \citep{papineni-etal-2002-bleu} and ROUGE \citep{lin-2004-rouge} that prioritize correct word prediction was not prioritized in our study. It is considered that previously discussed Word Coverage and Semantic Similarity metrics were better suited for evaluating the proposed framework.

The approach taken for condition structures is similar to pseudo-relevance feedback approaches. The differentiating aspect here is that obtaining a document or a list of documents that are likely to match the user query requires multiple operations within the search engine. As this requirement would hinder the time performance of the query-document matching, we avoided utilizing pseudo-relevance feedback approaches directly in our studies. Applied condition mechanisms are designed to be stored outside the search engine environment and memory. Hence, operations needed for reaching condition vectors are not reflected in the performance of the search engine. However, the time and space requirements of these conditions are the primary drawbacks of these approaches. As for all condition structures discussed construction of a lookup table is necessary, these lookup tables should be generated or updated before model training and tuning. Thus, these structures form an additional step for complete model training.

Applying the word and document similarity for conditions intends to enrich the initial user query that often consists of one to two words. However, we observed that with the word and document similarity condition mechanisms assistance for rare user queries may not be adequate to decrease the effects of the cold start problem. The reason for this is that the condition vectors obtained by word similarity and document similarity may affect the sentence production of the model in undesirable ways. The sequences that can be produced with the word and document information added with the conditions can be differentiated from the document information corresponding to the search made by the user. When a user query consisting of very few words is combined with conditions that are almost the same size and contain the same amount of semantic meaning, the indexes produced by the model can diverge from the sequences desired to be obtained.

There are differences between the conditioned GAN architectures in the literature and the conditioned GAN architectures presented in this study. It has been seen that the conditional GAN architectures in the literature utilize conditions for both the generator and the discriminator models. In these studies, it is a correct approach to feed both the generator and the discriminator with this information, as the conditions are usually made up of class labels. In our study, since the condition structures consist of semantic information that increases the sentence generation performance of the model, the condition structures were used only in the generators. The discriminator model only performs binary classification between synthetic data and product information corresponding to user queries. Another differentiating issue is the training phase of the generative model. In the \textit{m}QE-CGAN architecture, Monte Carlo simulations were not used in the pre-training phase of the generative models. Softmax operations were used in an iterative manner for the models to predict the next words in the sequences. Monte Carlo simulations were used only during adversarial learning.

It is observed that similar semantic evaluation metric values can be obtained with adversarial learning. For the dataset of Company 2, the adversarial learning phase improves the Semantic Similarity metric between generated and reference sequences from 0.911 to 0.914. Likewise, the Semantic Similarity metric increases from 0.736 to 0.808 for the dataset of Company 4. Hence, the average cosine similarity between generated sequences and reference documents increase by nearly 10\%  after the adversarial learning phase compared to the generator evaluation metrics with the baseline model. For other datasets, adversarial training until the generator loss function does not improve did not yield better semantic evaluation metrics. It suggests that there are further tuning and optimization steps for the adversarial learning process of the framework. Due to this, we take the 10\% performance increase as the best improvement of the adversarial learning phase. 

The table below displays the sequences obtained after the adversarial learning phase of the \textit{m}QE-CGAN framework. 

\begin{table*}[ht]
    \captionsetup{font=scriptsize}
    \centering
    \resizebox{1\linewidth}{!}{
    \begin{tabular}{ c|c|c|c } 
    \hline
    Dataset & Query & Generated Sequence & Reference Document \\
    \hline
    \hline
    \multirow{4}{4 em}{C1}  & saçlar & saçlar nemlendirici krem 50 & water nemlendirici şampuan  \\ 
    & köpük & karma köpük 150 & vitaminli 150 ml yüz köpüğü \\ 
    & yıpranma krem & yıpranmış nemlendirici krem 50 & {brand name} yıpranma karşıtı nemlendirici krem 50 ml \\ 
    \hline
    \multirow{4}{4 em}{C2}  & bandana & siyah parka & sarı bucket çanta  \\ 
    & krem ceket & kapüşonlu siyah ceket & kapüşonlu beyaz ceket \\ 
    & \{company name\} black jake & jake \{company name\} black jean pantolon & jake \{company name\} black gölgeli jean pantolon \\ 
    \hline
    \multirow{4}{4 em}{C3}  & şarj & \{company name\} \{model name\} c-type hızlı seyahat & \{company name\} \{model name\} siyah \\ 
    & \{company name\} \{model name\} & \{company name\} \{model name\} 128 gb & \{company name\} \{model name\} 128 gb \\ 
    & kulaklık & \{company name\} \{model name\} kablolu mikrofonlu & \{company name\} \{model name\} kulaklık \\ 
    \hline
    \multirow{4}{4 em}{C4}  & mont & \{company name\} klasik regular fit & standart fit mont  \\ 
    & kareli gömlek & slim fit gömlek & slim fit kareli gömlek \\ 
    & polo yaka tisort & \{company name\} polo yaka cepsiz & regular fit polo yaka tisort \\ 
    \hline
    \end{tabular}
    \caption{Randomly selected generated samples and their corresponding query and reference document pairs. Generated sequences are obtained after the adversarial learning The generator of the framework was selected as Word Similarity model. For each different company dataset, three examples are displayed in the table. Whenever the company name is included in the generated sequence, they are marked as \{company name\}. \{brand name\} is added to not reveal specific brand names in the C3 dataset. \{model name\} is added to hide specific product models in the C3 to not reveal the further information about the company.}
    \label{tab:seqs}}
\end{table*}

When examples in Table \ref{tab:seqs} are analyzed, it can be seen that the model tends to generate the company name as an expansion term. It is because company names are the most common words in the case of C2 and C4 datasets. Thus, the models have the bias of outputting the most occurred word in the dataset. For these datasets, the expansion terms seem to be meaningful in general. For user queries such as "mont" (coat) or "gömlek" (shirt), the model generates terms such as "regular fit" or "slim fit". On the other hand, when the initial user query does not have a matching document in the dataset, the generated expansion terms seem to be less successful. The most obvious example of this observation is the first example given for C2. The query "bandana" is expanded with "siyah parka" (black parka) where the query initially matches with the document "sarı bucket çanta" (yellow bucket bag).

It seems that trained models are more successful for the expansion generation task where the relationship between words is more precise. If the candidate words to expand the given query are more limited, models seem to capture the semantic relationship between different words in a smaller space. Generation results of the C3 dataset exemplify this phenomenon. In the first example, the query "şarj" (charge) is expanded with words such as "c-type", "hızlı" (fast), and "seyehat" (travel). The second example adds the memory information that is very common to be included in product names to the search query of a specific device. The third example adds "kablolu" (wired) and "mikrofonlu" (with microphone) to the user query of "kulaklık" (headphones). 

The presence of this phenomenon can also be seen in the results of the C1 dataset. The query "saçlar" (hair) is paired with "nemlendirici" (moisturizer) and "krem" (conditioner). In the third example, the query "yıpranma krem" is expanded with "nemlendirici" (moisturizer) and the correct volume of the product. As the C1 dataset is mostly composed of cosmetics products, the dataset usually consists of documents that have volume information. Results display that the trained model is not successful at generating sequences with correct volume information consisting of the volume value and its unit. We observed that our model tended to include a numerical value to generated sequences often but did not include its unit such as "ml" or "cc". It suggests that models can be further optimized to capture the relationships between individual word pairs in a better way.

\section{Conclusion}
Our work focused on bringing concepts of generative adversarial networks, query expansion, and condition structures originated from query-document relationships together. Results from the \textit{m}QE-CGAN framework demonstrate that given user queries with limited information can be enriched with query expansion to obtain sequences that are semantically more similar to the documents in the datasets. As the trained models yield successful evaluation metrics for capturing the context of given query-document pairs, utilization of the framework can be beneficial for optimizing search engines in the e-commerce domain.

Various aspects of the proposed GAN framework can be improved. Firstly, we believe that the sequence generation process could benefit from utilizing context-specific word embeddings. To this end, word embeddings obtained from language models fine-tuned for datasets will be tested in the future. Secondly, alternative condition mechanisms can be introduced during the training process. The proposed framework allows the replacement of condition mechanisms to adapt specific cases by capturing different semantic relationships in query-document data. One of the condition structures to be applied is the combination of the conditions experimented with in this study. Lastly, we aim to experiment with the integration of the proposed GAN framework with the existing search engine. This way, the advantages and shortcomings of a search engine with an integrated GAN model for query expansion can be observed in high-traffic environments. In future works, we aim to assess the practical evaluation metrics of the query expansion approach for its performance against the cold start problem.


\bibliographystyle{cas-model2-names}

\bibliography{cas-refs}

\begin{thebibliography}{28}
\expandafter\ifx\csname natexlab\endcsname\relax\def\natexlab#1{#1}\fi
\providecommand{\url}[1]{\texttt{#1}}
\providecommand{\href}[2]{#2}
\providecommand{\path}[1]{#1}
\providecommand{\DOIprefix}{doi:}
\providecommand{\ArXivprefix}{arXiv:}
\providecommand{\URLprefix}{URL: }
\providecommand{\Pubmedprefix}{pmid:}
\providecommand{\doi}[1]{\href{http://dx.doi.org/#1}{\path{#1}}}
\providecommand{\Pubmed}[1]{\href{pmid:#1}{\path{#1}}}
\providecommand{\bibinfo}[2]{#2}
\ifx\xfnm\relax \def\xfnm[#1]{\unskip,\space#1}\fi
\bibitem[{Azad and Deepak(2019a)}]{azadanew}
\bibinfo{author}{Azad, H.K.}, \bibinfo{author}{Deepak, A.},
  \bibinfo{year}{2019}a.
\newblock \bibinfo{title}{A new approach for query expansion using wikipedia
  and wordnet}.
\newblock \bibinfo{journal}{Information Sciences} \bibinfo{volume}{492},
  \bibinfo{pages}{147--163}.
\newblock \URLprefix
  \url{https://www.sciencedirect.com/science/article/pii/S0020025519303263},
  \DOIprefix\doi{https://doi.org/10.1016/j.ins.2019.04.019}.
\bibitem[{Azad and Deepak(2019b)}]{Azad_2019}
\bibinfo{author}{Azad, H.K.}, \bibinfo{author}{Deepak, A.},
  \bibinfo{year}{2019}b.
\newblock \bibinfo{title}{Query expansion techniques for information retrieval:
  A survey}.
\newblock \bibinfo{journal}{Information Processing and Management}
  \bibinfo{volume}{56}, \bibinfo{pages}{1698--1735}.
\newblock \URLprefix \url{https://doi.org/10.1016%2Fj.ipm.2019.05.009},
  \DOIprefix\doi{10.1016/j.ipm.2019.05.009}.
\bibitem[{Bojanowski et~al.(2016)Bojanowski, Grave, Joulin and
  Mikolov}]{bojanowski2016enriching}
\bibinfo{author}{Bojanowski, P.}, \bibinfo{author}{Grave, E.},
  \bibinfo{author}{Joulin, A.}, \bibinfo{author}{Mikolov, T.},
  \bibinfo{year}{2016}.
\newblock \bibinfo{title}{Enriching word vectors with subword information}.
\newblock \bibinfo{journal}{arXiv preprint arXiv:1607.04606} .
\bibitem[{Camacho and Alves-Souza(2018)}]{camacho2018social}
\bibinfo{author}{Camacho, L.A.G.}, \bibinfo{author}{Alves-Souza, S.N.},
  \bibinfo{year}{2018}.
\newblock \bibinfo{title}{Social network data to alleviate cold-start in
  recommender system: A systematic review}.
\newblock \bibinfo{journal}{Information Processing \& Management}
  \bibinfo{volume}{54}, \bibinfo{pages}{529--544}.
\bibitem[{Carpineto et~al.(2001)Carpineto, de~Mori, Romano and
  Bigi}]{carpineto2001}
\bibinfo{author}{Carpineto, C.}, \bibinfo{author}{de~Mori, R.},
  \bibinfo{author}{Romano, G.}, \bibinfo{author}{Bigi, B.},
  \bibinfo{year}{2001}.
\newblock \bibinfo{title}{An information-theoretic approach to automatic query
  expansion}.
\newblock \bibinfo{journal}{ACM Trans. Inf. Syst.} \bibinfo{volume}{19},
  \bibinfo{pages}{1–27}.
\newblock \URLprefix \url{https://doi.org/10.1145/366836.366860},
  \DOIprefix\doi{10.1145/366836.366860}.
\bibitem[{Carpineto and Romano(2012)}]{survey2012}
\bibinfo{author}{Carpineto, C.}, \bibinfo{author}{Romano, G.},
  \bibinfo{year}{2012}.
\newblock \bibinfo{title}{A survey of automatic query expansion in information
  retrieval}.
\newblock \bibinfo{journal}{ACM Comput. Surv.} \bibinfo{volume}{44},
  \bibinfo{pages}{1}.
\newblock \DOIprefix\doi{10.1145/2071389.2071390}.
\bibitem[{Diaz et~al.(2016)Diaz, Mitra and Craswell}]{localembed}
\bibinfo{author}{Diaz, F.}, \bibinfo{author}{Mitra, B.},
  \bibinfo{author}{Craswell, N.}, \bibinfo{year}{2016}.
\newblock \bibinfo{title}{Query expansion with locally-trained word
  embeddings}.
\newblock \URLprefix \url{https://arxiv.org/abs/1605.07891},
  \DOIprefix\doi{10.48550/ARXIV.1605.07891}.
\bibitem[{Furnas et~al.(1987)Furnas, Landauer, Gomez and Dumais}]{furnas}
\bibinfo{author}{Furnas, G.W.}, \bibinfo{author}{Landauer, T.K.},
  \bibinfo{author}{Gomez, L.M.}, \bibinfo{author}{Dumais, S.T.},
  \bibinfo{year}{1987}.
\newblock \bibinfo{title}{The vocabulary problem in human-system
  communication}.
\newblock \bibinfo{journal}{Commun. ACM} \bibinfo{volume}{30},
  \bibinfo{pages}{964–971}.
\newblock \URLprefix \url{https://doi.org/10.1145/32206.32212},
  \DOIprefix\doi{10.1145/32206.32212}.
\bibitem[{Huang et~al.(2021)Huang, Wang, Liu and Ding}]{huang2021gqe}
\bibinfo{author}{Huang, M.}, \bibinfo{author}{Wang, D.}, \bibinfo{author}{Liu,
  S.}, \bibinfo{author}{Ding, M.}, \bibinfo{year}{2021}.
\newblock \bibinfo{title}{Gqe-prf: Generative query expansion with
  pseudo-relevance feedback}.
\newblock \bibinfo{journal}{arXiv preprint arXiv:2108.06010} .
\bibitem[{Kingma and Ba(2014)}]{adam}
\bibinfo{author}{Kingma, D.P.}, \bibinfo{author}{Ba, J.}, \bibinfo{year}{2014}.
\newblock \bibinfo{title}{Adam: A method for stochastic optimization}.
\newblock \URLprefix \url{https://arxiv.org/abs/1412.6980},
  \DOIprefix\doi{10.48550/ARXIV.1412.6980}.
\bibitem[{Kusner and Hernández-Lobato(2016)}]{gumbel}
\bibinfo{author}{Kusner, M.J.}, \bibinfo{author}{Hernández-Lobato, J.M.},
  \bibinfo{year}{2016}.
\newblock \bibinfo{title}{Gans for sequences of discrete elements with the
  gumbel-softmax distribution}.
\newblock \URLprefix \url{https://arxiv.org/abs/1611.04051},
  \DOIprefix\doi{10.48550/ARXIV.1611.04051}.
\bibitem[{Lee et~al.(2018)Lee, Gao and Zhang}]{lee2018rare}
\bibinfo{author}{Lee, M.C.}, \bibinfo{author}{Gao, B.}, \bibinfo{author}{Zhang,
  R.}, \bibinfo{year}{2018}.
\newblock \bibinfo{title}{Rare query expansion through generative adversarial
  networks in search advertising}, in: \bibinfo{booktitle}{Proceedings of the
  24th acm sigkdd international conference on knowledge discovery \& data
  mining}, pp. \bibinfo{pages}{500--508}.
\bibitem[{Lian et~al.(2021)Lian, Chen, Jia, You, Tian, Hu, Zhang, Yan, Tong,
  Han et~al.}]{lian2021end}
\bibinfo{author}{Lian, Y.}, \bibinfo{author}{Chen, Z.}, \bibinfo{author}{Jia,
  J.}, \bibinfo{author}{You, Z.}, \bibinfo{author}{Tian, C.},
  \bibinfo{author}{Hu, J.}, \bibinfo{author}{Zhang, K.}, \bibinfo{author}{Yan,
  C.}, \bibinfo{author}{Tong, M.}, \bibinfo{author}{Han, W.}, et~al.,
  \bibinfo{year}{2021}.
\newblock \bibinfo{title}{An end-to-end generative retrieval method for
  sponsored search} .
\bibitem[{Lin(2004)}]{lin-2004-rouge}
\bibinfo{author}{Lin, C.Y.}, \bibinfo{year}{2004}.
\newblock \bibinfo{title}{{ROUGE}: A package for automatic evaluation of
  summaries}, in: \bibinfo{booktitle}{Text Summarization Branches Out},
  \bibinfo{publisher}{Association for Computational Linguistics},
  \bibinfo{address}{Barcelona, Spain}. pp. \bibinfo{pages}{74--81}.
\newblock \URLprefix \url{https://aclanthology.org/W04-1013}.
\bibitem[{Metzler and Croft(2007)}]{rm3}
\bibinfo{author}{Metzler, D.}, \bibinfo{author}{Croft, W.B.},
  \bibinfo{year}{2007}.
\newblock \bibinfo{title}{Latent concept expansion using markov random fields},
  in: \bibinfo{booktitle}{Proceedings of the 30th Annual International ACM
  SIGIR Conference on Research and Development in Information Retrieval},
  \bibinfo{publisher}{Association for Computing Machinery},
  \bibinfo{address}{New York, NY, USA}. p. \bibinfo{pages}{311–318}.
\newblock \URLprefix \url{https://doi.org/10.1145/1277741.1277796},
  \DOIprefix\doi{10.1145/1277741.1277796}.
\bibitem[{Mikolov et~al.(2013)Mikolov, Chen, Corrado and Dean}]{word2vec}
\bibinfo{author}{Mikolov, T.}, \bibinfo{author}{Chen, K.},
  \bibinfo{author}{Corrado, G.}, \bibinfo{author}{Dean, J.},
  \bibinfo{year}{2013}.
\newblock \bibinfo{title}{Efficient estimation of word representations in
  vector space}.
\newblock \URLprefix \url{https://arxiv.org/abs/1301.3781},
  \DOIprefix\doi{10.48550/ARXIV.1301.3781}.
\bibitem[{Mirza and Osindero(2014)}]{cgan}
\bibinfo{author}{Mirza, M.}, \bibinfo{author}{Osindero, S.},
  \bibinfo{year}{2014}.
\newblock \bibinfo{title}{Conditional generative adversarial nets}.
\newblock \URLprefix \url{https://arxiv.org/abs/1411.1784},
  \DOIprefix\doi{10.48550/ARXIV.1411.1784}.
\bibitem[{Pal et~al.(2015)Pal, Mitra and Bhattacharya}]{qucat}
\bibinfo{author}{Pal, D.}, \bibinfo{author}{Mitra, M.},
  \bibinfo{author}{Bhattacharya, S.}, \bibinfo{year}{2015}.
\newblock \bibinfo{title}{Exploring query categorisation for query expansion:
  {A} study}.
\newblock \bibinfo{journal}{CoRR} \bibinfo{volume}{abs/1509.05567}.
\newblock \URLprefix \url{http://arxiv.org/abs/1509.05567},
  \href{http://arxiv.org/abs/1509.05567}{\tt arXiv:1509.05567}.
\bibitem[{Papineni et~al.(2002)Papineni, Roukos, Ward and
  Zhu}]{papineni-etal-2002-bleu}
\bibinfo{author}{Papineni, K.}, \bibinfo{author}{Roukos, S.},
  \bibinfo{author}{Ward, T.}, \bibinfo{author}{Zhu, W.J.},
  \bibinfo{year}{2002}.
\newblock \bibinfo{title}{{B}leu: a method for automatic evaluation of machine
  translation}, in: \bibinfo{booktitle}{Proceedings of the 40th Annual Meeting
  of the Association for Computational Linguistics},
  \bibinfo{publisher}{Association for Computational Linguistics},
  \bibinfo{address}{Philadelphia, Pennsylvania, USA}. pp.
  \bibinfo{pages}{311--318}.
\newblock \URLprefix \url{https://aclanthology.org/P02-1040},
  \DOIprefix\doi{10.3115/1073083.1073135}.
\bibitem[{Paszke et~al.(2019)Paszke, Gross, Massa, Lerer, Bradbury, Chanan,
  Killeen, Lin, Gimelshein, Antiga, Desmaison, Kopf, Yang, DeVito, Raison,
  Tejani, Chilamkurthy, Steiner, Fang, Bai and Chintala}]{NEURIPS2019_9015}
\bibinfo{author}{Paszke, A.}, \bibinfo{author}{Gross, S.},
  \bibinfo{author}{Massa, F.}, \bibinfo{author}{Lerer, A.},
  \bibinfo{author}{Bradbury, J.}, \bibinfo{author}{Chanan, G.},
  \bibinfo{author}{Killeen, T.}, \bibinfo{author}{Lin, Z.},
  \bibinfo{author}{Gimelshein, N.}, \bibinfo{author}{Antiga, L.},
  \bibinfo{author}{Desmaison, A.}, \bibinfo{author}{Kopf, A.},
  \bibinfo{author}{Yang, E.}, \bibinfo{author}{DeVito, Z.},
  \bibinfo{author}{Raison, M.}, \bibinfo{author}{Tejani, A.},
  \bibinfo{author}{Chilamkurthy, S.}, \bibinfo{author}{Steiner, B.},
  \bibinfo{author}{Fang, L.}, \bibinfo{author}{Bai, J.},
  \bibinfo{author}{Chintala, S.}, \bibinfo{year}{2019}.
\newblock \bibinfo{title}{Pytorch: An imperative style, high-performance deep
  learning library}, in: \bibinfo{booktitle}{Advances in Neural Information
  Processing Systems 32}. \bibinfo{publisher}{Curran Associates, Inc.}, pp.
  \bibinfo{pages}{8024--8035}.
\newblock \URLprefix
  \url{http://papers.neurips.cc/paper/9015-pytorch-an-imperative-style-high-performance-
  deep-learning-library.pdf}.
\bibitem[{Qi et~al.(2020)Qi, Gong, Yan, Jiao, Shao, Zhang, Li, Duan and
  Zhou}]{prophetnetads}
\bibinfo{author}{Qi, W.}, \bibinfo{author}{Gong, Y.}, \bibinfo{author}{Yan,
  Y.}, \bibinfo{author}{Jiao, J.}, \bibinfo{author}{Shao, B.},
  \bibinfo{author}{Zhang, R.}, \bibinfo{author}{Li, H.}, \bibinfo{author}{Duan,
  N.}, \bibinfo{author}{Zhou, M.}, \bibinfo{year}{2020}.
\newblock \bibinfo{title}{Prophetnet-ads: A looking ahead strategy for
  generative retrieval models in sponsored search engine}, in:
  \bibinfo{editor}{Zhu, X.}, \bibinfo{editor}{Zhang, M.},
  \bibinfo{editor}{Hong, Y.}, \bibinfo{editor}{He, R.} (Eds.),
  \bibinfo{booktitle}{Natural Language Processing and Chinese Computing},
  \bibinfo{publisher}{Springer International Publishing},
  \bibinfo{address}{Cham}. pp. \bibinfo{pages}{305--317}.
\bibitem[{Sohn et~al.(2015)Sohn, Lee and Yan}]{NIPS2015_8d55a249}
\bibinfo{author}{Sohn, K.}, \bibinfo{author}{Lee, H.}, \bibinfo{author}{Yan,
  X.}, \bibinfo{year}{2015}.
\newblock \bibinfo{title}{Learning structured output representation using deep
  conditional generative models}, in: \bibinfo{editor}{Cortes, C.},
  \bibinfo{editor}{Lawrence, N.}, \bibinfo{editor}{Lee, D.},
  \bibinfo{editor}{Sugiyama, M.}, \bibinfo{editor}{Garnett, R.} (Eds.),
  \bibinfo{booktitle}{Advances in Neural Information Processing Systems},
  \bibinfo{publisher}{Curran Associates, Inc.}
\newblock \URLprefix
  \url{https://proceedings.neurips.cc/paper/2015/file/8d55a249e6baa5c06772297520da2051-Paper.pdf}.
\bibitem[{Sordoni et~al.(2014)Sordoni, Bengio and
  Nie}]{Sordoni_Bengio_Nie_2014}
\bibinfo{author}{Sordoni, A.}, \bibinfo{author}{Bengio, Y.},
  \bibinfo{author}{Nie, J.Y.}, \bibinfo{year}{2014}.
\newblock \bibinfo{title}{Learning concept embeddings for query expansion by
  quantum entropy minimization}.
\newblock \bibinfo{journal}{Proceedings of the AAAI Conference on Artificial
  Intelligence} \bibinfo{volume}{28}.
\newblock \URLprefix
  \url{https://ojs.aaai.org/index.php/AAAI/article/view/8933},
  \DOIprefix\doi{10.1609/aaai.v28i1.8933}.
\bibitem[{Spink et~al.(2001)Spink, Wolfram, Jansen and Saracevic}]{powerlaw}
\bibinfo{author}{Spink, A.}, \bibinfo{author}{Wolfram, D.},
  \bibinfo{author}{Jansen, M.B.J.}, \bibinfo{author}{Saracevic, T.},
  \bibinfo{year}{2001}.
\newblock \bibinfo{title}{Searching the web: The public and their queries}.
\newblock \bibinfo{journal}{Journal of the American Society for Information
  Science and Technology} \bibinfo{volume}{52}, \bibinfo{pages}{226--234}.
\newblock
  \DOIprefix\doi{https://doi.org/10.1002/1097-4571(2000)9999:9999<::AID-ASI1591>3.0.CO;2-R}.
\bibitem[{Sutton et~al.(1999)Sutton, McAllester, Singh and
  Mansour}]{policy_grad}
\bibinfo{author}{Sutton, R.S.}, \bibinfo{author}{McAllester, D.},
  \bibinfo{author}{Singh, S.}, \bibinfo{author}{Mansour, Y.},
  \bibinfo{year}{1999}.
\newblock \bibinfo{title}{Policy gradient methods for reinforcement learning
  with function approximation}, in: \bibinfo{editor}{Solla, S.},
  \bibinfo{editor}{Leen, T.}, \bibinfo{editor}{M\"{u}ller, K.} (Eds.),
  \bibinfo{booktitle}{Advances in Neural Information Processing Systems},
  \bibinfo{publisher}{MIT Press}.
\newblock \URLprefix
  \url{https://proceedings.neurips.cc/paper/1999/file/464d828b85b0bed98e80ade0a5c43b0f-Paper.pdf}.
\bibitem[{Symonds et~al.(2011)Symonds, Bruza, Sitbon and Turner}]{tensorr}
\bibinfo{author}{Symonds, M.}, \bibinfo{author}{Bruza, P.},
  \bibinfo{author}{Sitbon, L.}, \bibinfo{author}{Turner, I.},
  \bibinfo{year}{2011}.
\newblock \bibinfo{title}{Tensor query expansion: A cognitively motivated
  relevance model}.
\bibitem[{Vaswani et~al.(2017)Vaswani, Shazeer, Parmar, Uszkoreit, Jones,
  Gomez, Kaiser and Polosukhin}]{NIPS2017_3f5ee243}
\bibinfo{author}{Vaswani, A.}, \bibinfo{author}{Shazeer, N.},
  \bibinfo{author}{Parmar, N.}, \bibinfo{author}{Uszkoreit, J.},
  \bibinfo{author}{Jones, L.}, \bibinfo{author}{Gomez, A.N.},
  \bibinfo{author}{Kaiser, L.u.}, \bibinfo{author}{Polosukhin, I.},
  \bibinfo{year}{2017}.
\newblock \bibinfo{title}{Attention is all you need}, in:
  \bibinfo{editor}{Guyon, I.}, \bibinfo{editor}{Luxburg, U.V.},
  \bibinfo{editor}{Bengio, S.}, \bibinfo{editor}{Wallach, H.},
  \bibinfo{editor}{Fergus, R.}, \bibinfo{editor}{Vishwanathan, S.},
  \bibinfo{editor}{Garnett, R.} (Eds.), \bibinfo{booktitle}{Advances in Neural
  Information Processing Systems}, \bibinfo{publisher}{Curran Associates, Inc.}
\newblock \URLprefix
  \url{https://proceedings.neurips.cc/paper/2017/file/3f5ee243547dee91fbd053c1c4a845aa-Paper.pdf}.
\bibitem[{Yu et~al.(2017)Yu, Zhang, Wang and Yu}]{yu2017seqgan}
\bibinfo{author}{Yu, L.}, \bibinfo{author}{Zhang, W.}, \bibinfo{author}{Wang,
  J.}, \bibinfo{author}{Yu, Y.}, \bibinfo{year}{2017}.
\newblock \bibinfo{title}{Seqgan: Sequence generative adversarial nets with
  policy gradient}, in: \bibinfo{booktitle}{Proceedings of the AAAI conference
  on artificial intelligence}.

\end{thebibliography}

\end{document}